\documentclass[a4paper]{article}
\usepackage{INTERSPEECH2021}

\usepackage{amsmath}
\usepackage{graphicx}

\usepackage{flushend}
\usepackage[colorlinks=true, allcolors=blue]{hyperref}
\usepackage{cite}
\usepackage{booktabs} 
\usepackage{balance}
\usepackage{courier}  
\usepackage{url}  
\usepackage[ruled,lined,vlined,linesnumbered]{algorithm2e}
\usepackage[noend]{algorithmic}
\usepackage{bm}
\usepackage{url}
\usepackage{enumitem}
\usepackage{changepage}
\usepackage{color}
\usepackage{multirow}
\usepackage{amsfonts} 
\usepackage{amssymb}
\usepackage{float}
\usepackage{soul}
\usepackage{subfig}
\usepackage{cleveref}
\usepackage{bbm}
\usepackage{subfig}
\usepackage{footnote}
\makesavenoteenv{tabular}

\def\x{{\mathbf x}}

\newcommand{\rr}[1]{{\color{blue}[Ruirui: #1]}}

\def\model{\textsc{foe}net\xspace}

\providecommand{\keywords}[1]
{
  \small	
  \textbf{\textit{Keywords---}} 
}

\title{Fusion of Embeddings Networks for
Robust Combination of \\
Text Dependent and Independent Speaker Recognition}
\name{Ruirui Li \quad Chelsea J.-T. Ju \quad Zeya Chen \quad Hongda Mao \quad Oguz Elibol \quad Andreas Stolcke}
\address{Amazon Alexa Speech} 
\email{
	\{ruirul, juitij, zeyachen, hongdam, oelibol, stolcke\}@amazon.com}

\begin{document}
\ninept
\maketitle
\begin{abstract}
By implicitly recognizing a user based on his/her speech input, speaker identification enables many downstream applications, such as personalized system behavior and expedited shopping checkouts.
Based on whether the speech content is constrained or not, both text-dependent (TD) and text-independent (TI) speaker recognition models may be used. 
We wish to combine the advantages of both types of models through an ensemble system to make more reliable predictions.
However, any such combined approach has to be robust to incomplete inputs, i.e., when either TD or TI input is missing.
As a solution we propose a \underline{f}usion \underline{o}f \underline{e}mbeddings \underline{net}work (\model) architecture, 
combining joint learning with neural attention.
We compare \model\ with four competitive baseline methods on a dataset of voice assistant inputs,
and show that it achieves higher accuracy than the baseline and score fusion methods, especially in the presence of incomplete inputs.
\end{abstract}

\section{Introduction}
\label{sec:intro}
Speaker recognition answers the question ``who is speaking'' given a speech utterance. Based on whether the utterance content is constrained or not, there are two types of speaker recognition model:  text-dependent (TD) and text-independent (TI). 
For the former, the utterance content is known or limited to a small set of options.
For deployment in voice assistants, the input usually consists of the predefined wakeword, such as ``Hey Google'', ``Echo'' or ``Alexa''.
TI systems, on the other hand, make no assumptions on what was said, and therefore have wider
applicability.
Besides, they are typically more accurate than TD systems as the utterances from TI systems tend to be longer than the ones from TD systems at both training and test time~\cite{DBLP:journals/corr/abs-2104-02125}.

For example, given an utterance like ``Alexa, what time is it?'', the TD and TI speaker recognition models can predict speaker identity independently. 
The TD model compares the acoustic signal corresponding to ``Alexa'' and a user's corresponding enrolled wakeword segments to calculate a matching score.
The TI model, on the other hand, can take either the entire utterance or the non-wakeword portion and compare it with users' enrolled utterances to yield another matching score.
The two models are generally trained with independent machine learning models and on different datasets with the goal of optimizing their recognition performance individually.
When both TD and TI models are available, we should be able to improve overall accuracy by 
combining their predictions; however, the following three concerns arise.
First, it is not clear how to best combine the models, given that a range of fusion methods have been proposed in machine learning (such as early fusion and late fusion~\cite{DBLP:conf/smc/GunesP05,DBLP:conf/mm/SnoekWS05}, etc.).
Second, the inputs for the two systems, especially the TD system, may not be always available.
For example, the wakeword may be optional in certain contexts. Table~\ref{table:senarios} further summarizes the scenarios with missing inputs from either the TD or the TI system.
Third, a user's voice may change (e.g., due to aging) and they may re-enroll, which may lead to a domain shift between training and test data, which in turn may affect the balance between TI and TD models for fusion purposes. 

\begin{table*}[tb]
\centering
\caption{Scenarios when speaker profile or utterance embeddings are unavailable.
$E_{spk}$ and $E_{u}$ represent speaker profile and test utterance embedding, respectively.
Detailed descriptions of $E_{spk}^{TD}$ and $E_{u}^{TD}$,$E_{spk}^{TI}$ and $E_{u}^{TI}$ are given in Section~\ref{subsec:emb_fusion}.}
\label{table:senarios}
\resizebox{0.90\linewidth}{!}{
\begin{tabular}{l||c|c|c|c} \hline
Scenario & $E_{spk}^{TD}$ & $E_{u}^{TD}$ & $E_{spk}^{TI}$& $E_{u}^{TI}$ \\ \hline \hline
There are no wakewords in an utterance & $\checkmark$ & $\x$ & $\checkmark$ & $\checkmark$ \\ \hline
The wakeword detector fails to identify the wakeword in an utterance & $\checkmark$ & $\x$ & $\checkmark$ & $\checkmark$ \\ \hline
Devices are woken up by a button press & $\checkmark$ & $\x$ & $\checkmark$ & $\checkmark$ \\ \hline
The speaker gets enrolled, but for a different wakeword & $\x$ & $\checkmark$ & $\checkmark$ & $\checkmark$ \\ \hline
TD model is working improperly during run time & $\checkmark$ & $\x$ & $\checkmark$ & $\checkmark$ \\ \hline
TI model is working improperly during run time & $\checkmark$ & $\checkmark$ & $\checkmark$ & $\x$ \\ \hline
\end{tabular}
}
\end{table*}

Speaker recognition has been studied extensively as a supervised classification problem~\cite{DBLP:conf/wsdm/LiJLH020, DBLP:conf/icassp/LiJWMH020, DBLP:conf/interspeech/LiJWHS20}. 
Traditionally, solutions are based on i-vector representation of speech segments~\cite{DBLP:journals/taslp/DehakKDDO11}.
More recently, methods based on deep neural net embeddings (d-vectors) have been adopted.
Deep speaker~\cite{DBLP:journals/corr/LiMJLZLCKZ17} and VGGVox~\cite{DBLP:journals/corr/abs-1806-05622} utilize CNN-based residual networks to extract audio embeddings and optimize triplet and contrastive losses to train speaker recognition models, respectively.
Two other Resnet-based methods~\cite{DBLP:conf/icassp/XieNCZ19, DBLP:journals/corr/abs-1807-08312} adopt additive margin softmax classification loss~\cite{DBLP:conf/iclr/0015LDL018} to improve the recognition accuracy.
More recently, Mockingjay~\cite{DBLP:conf/icassp/LiuYCHL20}, TERA~\cite{DBLP:journals/corr/abs-2007-06028}, APC~\cite{DBLP:conf/interspeech/ChungHTG19} and DeCoAR~\cite{DBLP:conf/icassp/LingLSK20} learn speaker embedding representations based on self-supervised spectrogram reconstructions or predictions.

While combining of speaker recognition models was very popular in research systems before d-vectors \cite{KajarekarEtAl:icassp2009,burget2009but}, little recent work has looked at combining multiple neural embedding systems.
\cite{DBLP:conf/icassp/ShonOG19} explores person verification by fusing audio and visual systems.
Researchers investigate how to fuse GMM-UBM based TD and TI speaker recognition models in~\cite{DBLP:conf/specom/MporasSS16}, where the TD and TI scores are fed as inputs to a fusion classifier. Nevertheless, it does not address the problem of incomplete inputs.

Our work assumes users have registered profiles consisting of embeddings, each based on four to ten utterances. We also assume pre-trained TD and TI models, which are capable of generating corresponding fix-length embeddings given a new utterance.
Unlike the aforementioned fusion research, which combines {\em score-level} outputs from component systems, we develop a fusion approach that uses {\em embedding-level} predictions from both TD and TI models.
Moreover, we allow the system to operate in the face of incomplete input, such as when an utterance does not start with a wakeword; the wakeword detector fails to identify the wakeword; a user has not yet enrolled for a wakeword sample (but the test utterance contains one); or an unexpected run-time error prevents the recognition engines from generating either TD or TI embeddings.

\begin{figure*}[tb]
    \centering
    \includegraphics[width=0.8\textwidth]{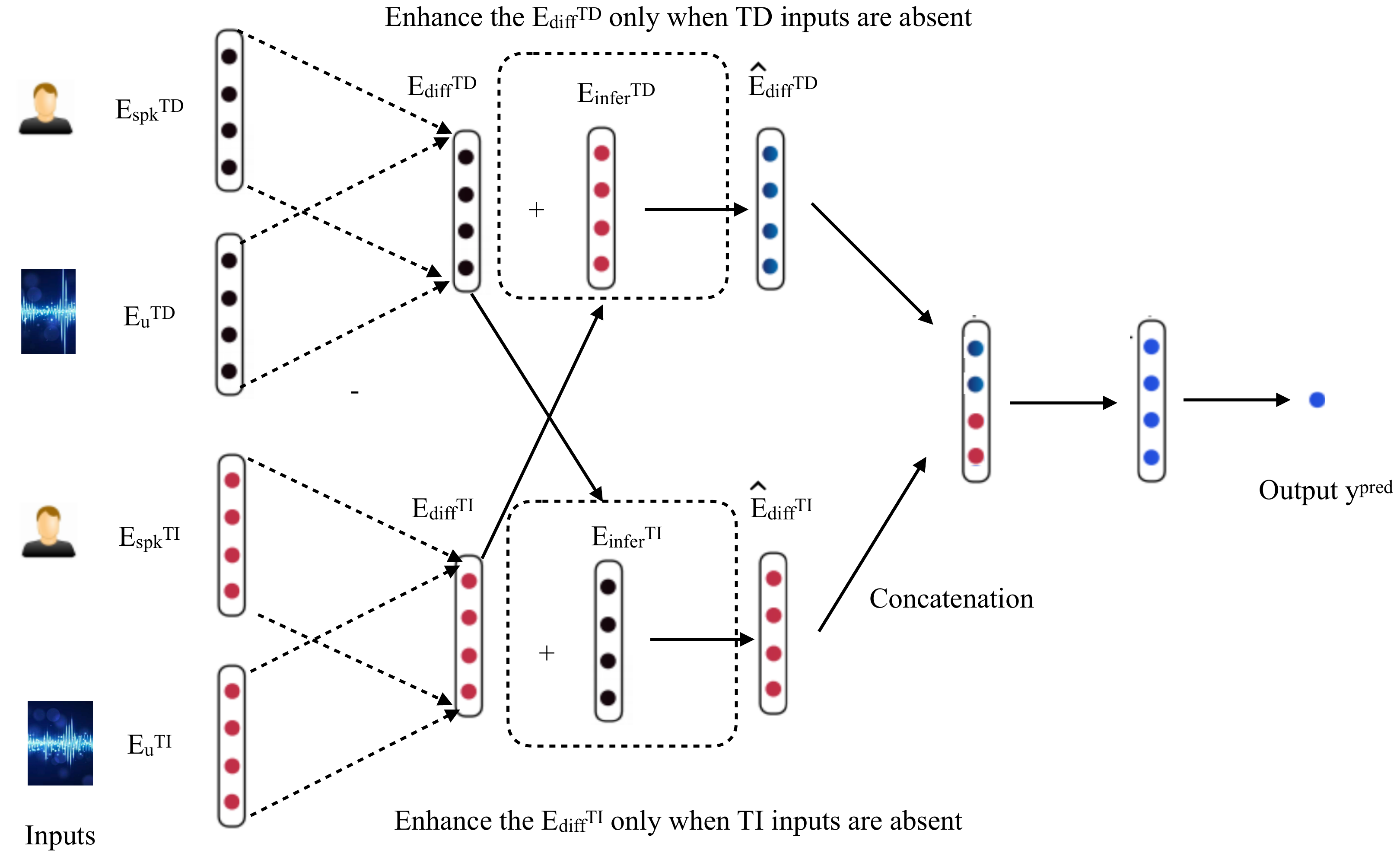}
    \caption{\emph{\model} architecture. Two differential embeddings $E_{\text{diff}}^{TD}$ and $E_{\text{diff}}^{TI}$ are first calculated based on speaker profile and test utterance embeddings,
    for TD and TI systems separately. In addition, two backup differential embeddings $E_{\text{infer}}^{TD}$ and $E_{\text{infer}}^{TI}$ are inferred from TI and TD systems, respectively.
    The backup differential embeddings are used only when $E_{\text{diff}}^{TD}$ or $E_{\text{diff}}^{TI}$, respectively, are not available. $\hat{E}_{diff}^{TD}$ and $\hat{E}_{diff}^{TI}$ combines the information from direct differential embedding and inferred differential embedding.
    Finally $\hat{E}_{diff}^{TD}$ and $\hat{E}_{diff}^{TI}$ are concatenated and fed to a connected layer.}
    \label{fig:framework}
\end{figure*}

\section{Fusion of Embeddings Network}
\label{subsec:emb_fusion}

Given a test utterance $u$, the goal of the speaker recognition model is to learn a scoring function such that the score $s(spk, u)$ between the target speaker and $u$ is as high as possible,
while the scores between other speakers and $u$ are as low as possible.

We use $E_{u}^{TD}$ and $E_{u}^{TI}$ to denote the embeddings of a test utterance $u$ generated by TD and TI models, respectively.
Similarly, we use $E_{spk}^{TD}$ and $E_{spk}^{TI}$ to denote the profile embeddings of speaker $spk$ generated by the TD and TI model, respectively.
More precisely, if the profile of a speaker is composed of a set of $M$ utterances $\{u_1, u_2, ..., u_M\}$, then $E_{spk} = \frac{1}{M} \sum_{i=1}^{M} E_{u_i}$.

To effectively fuse predictions from both TD and TI models, while handling the issue of incomplete inputs, we propose the fusion model shown in Fig.~\ref{fig:framework}, which combines TD and TI predictions at the embedding level.
When the input from one of the systems is not available, the model will {\em infer} the missing embedding from the one that is available.
This results in a consistent input to the fusion layer, which computes a scalar score from the concatenation of TD and TI embeddings-level outputs. 

For both TD and TI models, we first calculate a differential embedding between the speaker profile embedding $E_{spk}$ and the utterance embedding $E_{u}$.
Formally,
\begin{equation}
E_{diff}^{TD} =E_{spk}^{TD} - E_{u}^{TD},
\label{equ:td_diff}
\end{equation}
and
\begin{equation}
E_{diff}^{TI} =E_{spk}^{TI} - E_{u}^{TI},
\label{equ:ti_diff}
\end{equation}
where the operator $-$ denotes element-wise subtraction. 
Ideally, if the utterance $u$ comes from the target speaker $spk$, small values, which are close to a vector of zeros, are expected to be observed in both $E_{diff}^{TD}$ and $E_{diff}^{TI}$. This is because $E_{spk}$, a representative of $spk$'s multiple enrolled utterances, is expected to be close in the latent space to the embedding representation of the speaker's test utterance $u$.
If $u$ comes from a non-target speaker, on the other hand, values very different from zeros are expected in both $E_{diff}^{TD}$ and $E_{diff}^{TI}$.
We posit that both $E_{diff}^{TD}$ and $E_{diff}^{TI}$ serve as useful discriminants, which help effectively distinguish genuine speakers from others and expedite model training.
Moreover, we expect that $E_{diff}^{TD}$ and $E_{diff}^{TI}$ remain discriminative as users' profiles and test utterances evolve over time.
As users' voices evolve over time, their profiles can be continuously updated by adding their recent voice utterances, which can be achieved by comparing the recent voice interactions received from devices and users' profiles at that moment.
In this way, relative small values, close to zeros, remain expected in $E_{diff}^{TD}$ and $E_{diff}^{TI}$ if test utterances come from target speakers.

As detailed in Table~\ref{table:senarios}, $E_{spk}$ and $E_{u}$ are not always available. In such cases, we cannot calculate the differential embedding directly.
To address this issue, when either one or both of the two embedding inputs (i.e., $E_{spk}$ and $E_{u}$) from one system are not available, both $E_{spk}$ and $E_{u}$ from that system will be set to zero vectors. $E_{diff}$ will also be a zero vector as the result of element-wise subtraction.
At the same time, we will infer the differential embedding from the other subsystem, that is available.
Formally,
\begin{small}
\begin{equation}
\label{equ:td_diff_infer}
E_{infer}^{TD}=\begin{cases}
               0's, \text{if both $E_{spk}^{TD}$ and $E_{u}^{TD}$ are available,}\\
               f(E_{diff}^{TI} \cdot W_{TD2TI} + b_{TD2TI}), \text{Otherwise.}
            \end{cases}
\end{equation}
\end{small}
Similarly,
\begin{small}
\begin{equation}
\label{equ:ti_diff_infer}
E_{infer}^{TI}=\begin{cases}
               0's, \text{if both $E_{spk}^{TI}$ and $E_{u}^{TI}$ are available,}\\
               f(E_{diff}^{TD} \cdot W_{TI2TD} + b_{TI2TD}), \text{Otherwise.}
            \end{cases}
\end{equation}
\end{small}
where $W_{TD2TI}$, $W_{TI2TD}$ and $b_{TD2TI}$, $b_{TI2TD}$ are learnable weight matrices and bias vectors, respectively. $f(\cdot)$ is the activation function. 
Exponential linear units~\cite{DBLP:journals/corr/ClevertUH15} (ELUs) are chosen as the activation function for their expected benefits in convergence and accuracy.
The inferred differential embedding serves as a substitute that we only rely on when the direct differential embedding is not available.
Therefore,
\begin{equation}
\label{equ:td_enhanced_diff}
\hat{E}_{diff}^{TD}= E_{diff}^{TD} + E_{infer}^{TD}.
\end{equation}
Similarly,
\begin{equation}
\label{equ:ti_enhanced_diff}
\hat{E}_{diff}^{TI}= E_{diff}^{TI} + E_{infer}^{TI}.
\end{equation}

We then combine the differential information from TD and TI systems by concatenating $\hat{E}_{diff}^{TD}$ and $\hat{E}_{diff}^{TI}$.
\begin{equation}
\label{equ:combine_d_ti}
\hat{E}_{diff}= \hat{E}_{diff}^{TD} \| \hat{E}_{diff}^{TI}.
\end{equation}

We further feed $\hat{E}_{diff}$ to a fully connected layer to yield the prediction score $y^{pred}$. 
Formally,
\begin{equation}
\label{equ:prediction}
y^{pred}= f(BN(\hat{E}_{diff} \cdot W_{pred} + b_{pred})),
\end{equation}
where $W_{pred}$ and $b_{pred}$ are learnable parameters. $BN(\cdot)$ conducts the batch normalizations~\cite{DBLP:conf/icml/IoffeS15}, and $f(\cdot)$ is the activation function (the sigmoid function in our implementation).

We use cross-entropy as the loss for an instance $i$:
\begin{equation}
\label{eqn:adv_cls_loss_function}
l_i(\Theta) = -[y_i \log(y^{\text{pred}}_{i}) + (1 - y_i) \log(1-y^{\text{pred}}_{i})],
\end{equation}
where $\Theta$ denotes the union of all model parameters.

Given a mini-batch of $k$ training instances, the loss over all $k$ training instances is given by
\begin{equation}
\label{equ:entire_loss}
L(\Theta) = \frac{1}{k}\sum_{i = 1}^{k} l_i(\Theta) + \alpha L_2(\Theta),
\end{equation}
where $L_2(\Theta)$ calculates the $L_2$ norm of $\Theta$ as the regularizer and $\alpha$ is a hyperparameter, which balances the trade-off between the classification loss and regularizer. 
To update the model parameters $\Theta$, we apply Adam~\cite{DBLP:journals/corr/KingmaB14}.
Algorithm~\ref{alg:adaption} summarizes the detailed training processes of \emph{\model}.

\begin{algorithm}[th]\small
\caption{Embedding fusion model training}
\label{alg:adaption}
\textbf {Input:} learning rate $\eta$, maximal number of epochs $itr_{max}$, TD speaker embedding $E_{spk}^{TD}$, TD utterance embedding $E_{u}^{TD}$, TI speaker embedding $E_{spk}^{TI}$, TI utterance embedding $E_{u}^{TI}$\\
\textbf {Initialize:} Model parameters $\Theta$, EER$^*$ = $+\infty$
\begin{algorithmic}[1]
\STATE {/* Training on the training dataset */}
\FOR{$itr~\leq~itr_{max}$}
\STATE Randomly shuffle all training instances.
\STATE Sequentially generate a mini-batch of $k$ training instances.
\STATE Calculate loss based on Equation~\ref{equ:entire_loss}.
\STATE Update model parameters by applying Adam.
\STATE 
\STATE/* Calculate the EER on the validation dataset */
\IF {EER $\leq$ EER$^*$}
\STATE $\Theta^*$ = $\Theta$
\STATE EER$^*$ = EER
\ENDIF
\ENDFOR
\end{algorithmic}
\textbf {Output:} $\Theta^*$
\end{algorithm}

\section{Experiments}
\label{sec:experiments}
We follow previous work~\cite{DBLP:journals/corr/abs-1904-08775} to extract acoustic features from the raw audio.
Forty-dimensional Mel-spectrograms are extracted from waveforms after energy-based voice activity detection.

\subsection{Training and evaluation data}
\label{sec:training_eval_data}

The training and evaluation are conducted on de-identified speech utterances collected from Alexa devices.
The training data comprises both positive and negative instances.
To construct positive instances, we apply an existing speaker recognition model to 5,000-hour Alexa data and construct utterance/speaker pairs, as identified by high prediction scores in single-speaker registered households. In this way, it is less likely that non-target speakers are included in the positive training instances.
The same number of negative instances are then constructed by pairing a random guest speaker and a test utterance from the same gender but from a different household. The gender is determined by a pre-trained utterance-level gender classifier.
This selection strategy is based on the intuition that confusable (same-gender) pairs
are more informative for training.
After constructing the dataset, 85\% of instances are utilized as parameter training data and the remaining 15\%  serve as validation data, i.e., for model selection.

The evaluation dataset is constructed by first randomly sampling anonymized utterances. 
Then each sampled utterance and the enrollment data of speakers associated with the same device are sent to multiple annotators to collect the ground-truth labels independently.
To reduce annotation errors, we select utterances which have consistent annotation labels to form the final evaluation dataset.
For comparisons over time, we construct two such evaluation datasets based on data from two consecutive weeks.

\subsection{Baselines}

To evaluate the performance of \emph{\model}, the following four methods are adopted as baselines:
\begin{description}[leftmargin=0.5cm]
\item[\textbf{Single-system GE2E}]~\cite{Wan2018} utilizes an LSTM to construct utterance embeddings and optimizes the speaker recognition system by maximizing the similarity among utterances coming from the same speaker. It utilizes the entire utterance to make predictions and serves as a TI model. No TD system is used.

\item[\textbf{Average fusion (AF)}] takes the prediction scores from TD and TI models as inputs and outputs the average of these two scores as the fused prediction score.
When either TD or TI score is absent, we construct a piecewise linear score mapping based on pairs of corresponding FAR numbers between AF and the TD/TI system.

\item[\textbf{Score fusion (SF)}]~\cite{DBLP:conf/specom/MporasSS16} takes the prediction scores from the TD and TI systems as inputs and trains a neural network to learn the joint prediction. When the score from one system is absent, -1 is used as a placeholder input during both training and inference. 

\item[\textbf{Enhanced score fusion (E-SF)}] extends SF by inferring the score from the other SID system output when one score is absent.
The missing scores are estimated by nonlinear regression (linear transform followed by tanh) with the score from the other system as input,
and take the place of the missing fusion inputs in the SF network (see above).
\end{description}
We train the TD model based on the Siamese network~\cite{DBLP:conf/nips/BromleyGLSS93} and the TI model using GE2E~\cite{Wan2018}.
However, note that the proposed embedding fusion framework can be applied to any embedding-based TD and TI systems, or even to any sets of such systems where some of the outputs are sometimes unavailable.


\subsection{Results}
\label{sec:performance}
We evaluated the performance of \model\ against different baseline methods on the two one-week online evaluation datasets.
We calculate false accept rate (FAR) and false reject rate (FRR) on the evaluation sets.
\begin{equation}
\text{FAR} = \frac{\# \text{ of false accepted imposter trials}}{\# \text{ of imposter trials}}
\label{equ:far}
\end{equation}
\begin{equation}
\text{FRR} = \frac{\# \text{ of false rejected target trials}}{\# \text{ of target trials}}
\label{equ:frr}
\end{equation}
As different downstream applications have different requirements for expected FARs, we threshold scores at four different values, and compare corresponding FRR results across methods. 

\begin{table}[tb]
\centering
\caption{FRR relative improvements (\%) in Week 1}
\label{table:FRR_week_one}
\resizebox{1.0\linewidth}{!}{
\begin{tabular}{c|l|r|r|r|r} \hline
Scenarios & \multicolumn{1}{c}{Methods} & \multicolumn{4}{|c}{Targeted FAR} \\
            &       & 0.8\% & 2.0\% & 5.0\% & 12.5\% \\ \hline \hline
\multirow{4}{*}{TD\&TI present} &\emph{\model} vs GE2E & 21.0 & 22.5 & 22.8 & 22.5 \\ \cline{2-6}
& \emph{\model} vs AF & 10.3 & 10.3 & 14.4 & 14.1 \\ \cline{2-6}
& \emph{\model} vs SF & 11.9 & 13.2 & 15.0 & 13.8 \\ \cline{2-6}
& \emph{\model} vs E-SF & 11.3 & 12.1 & 14.9 & 13.8 \\ \hline\hline

\multirow{4}{*}{TD absent} &\emph{\model} vs GE2E & 14.8 & 19.4 & 30.5 & 47.7 \\ \cline{2-6}
& \emph{\model} vs AF & 17.2 & 20.3 & 31.7 & 49.3 \\ \cline{2-6}
& \emph{\model} vs SF & 6.0 & 14.4 & 29.1 & 48.3 \\ \cline{2-6}
& \emph{\model} vs E-SF & 4.2 & 13.3 & 27.6 & 47.0 \\ \hline\hline

\multirow{4}{*}{TI absent} &\emph{\model} vs GE2E & \multicolumn{4}{c}{\em not applicable} \\ \cline{2-6}
& \emph{\model} vs AF & 35.3 & 40.7 & 48.9 & 50.1 \\ \cline{2-6}
& \emph{\model} vs SF & 4.5 & 13.0 & 20.3 & 29.8 \\ \cline{2-6}
& \emph{\model} vs E-SF & 8.6 & 16.5 & 22.3 & 31.2 \\ \hline
\end{tabular}
}
\end{table}
\begin{table}[tb]
\centering
\caption{FRR relative improvements (\%) in Week 2}
\label{table:FRR_week_two}
\resizebox{1.0\linewidth}{!}{
\begin{tabular}{c|l|r|r|r|r} \hline
Scenarios & \multicolumn{1}{c}{Methods} & \multicolumn{4}{|c}{Targeted FAR} \\
            &       & 0.8\% & 2.0\% & 5.0\% & 12.5\% \\ \hline \hline
\multirow{3}{*}{TD\&TI present} &\emph{\model} vs GE2E & 20.5 & 21.7 & 25.0 & 22.5 \\ \cline{2-6}
& \emph{\model} vs AF & 8.3 & 12.1 & 13.1 & 14.6 \\ \cline{2-6}
& \emph{\model} vs SF & 12.1 & 14.3 & 14.3 & 16.5 \\ \cline{2-6}
& \emph{\model} vs E-SF & 10.5 & 13.3 & 13.7 & 17.4 \\ \hline\hline

\multirow{3}{*}{TD absent} &\emph{\model} vs GE2E & 14.3 & 20.6 & 30.7 & 45.0 \\ \cline{2-6}
& \emph{\model} vs AF & 15.2 & 20.6 & 30.7 & 46.0 \\ \cline{2-6}
& \emph{\model} vs SF & 5.2 & 14.3 & 28.1 & 45.2 \\ \cline{2-6}
& \emph{\model} vs E-SF & 4.0 & 13.0 & 26.6 & 43.8 \\ \hline\hline

\multirow{3}{*}{TI absent} &\emph{\model} vs GE2E & \multicolumn{4}{c}{\em not applicable} \\ \cline{2-6}
& \emph{\model} vs AF & 34.9 & 40.7 & 49.9 & 53.3 \\ \cline{2-6}
& \emph{\model} vs SF & 3.8 & 10.7 & 23.1 & 31.2 \\ \cline{2-6}
& \emph{\model} vs E-SF & 10.4 & 15.7 & 25.8 & 32.8 \\ \hline
\end{tabular}
}
\end{table}


Tables~\ref{table:FRR_week_one} and~\ref{table:FRR_week_two} compare the FRRs for different methods on the two evaluation datasets.
The tabulated values are the relative FRR reductions between \model and each of the four baselines.
We highlight four main observations. 
First, we observe that \model, AF, SF, and E-SF outperform single-system GE2E in most cases, especially when both TD and TI inputs are available. This shows the advantage of utilizing both TD and TI models in combination. Clearly, leveraging both TD and TI systems yields more accurate speaker identity predictions than just depending on a single TI model.

Second, \model\ consistently outperforms the fusion-by-average (AF) and the two score-fusion baselines (SF and E-SF).
For example, when we target 12.5\% FAR, compared with AF/SF/E-SF, \model\ reduces FRR by 14.1\%/13.8\%/13.8\%, respectively, when both TD and TI inputs are available in Week~1. We observe similar FRR reductions when targeting 0.8\%, 2.0\%, and 5.0\% FARs.
The same observation applies to the Week-2 evaluation set.
We conclude that fusion at the embedding level, rather than at the score level, significantly improves the speaker recognition performance.

Third, for almost all FAR levels, the advantage of \model\ over GE2E, AF, SF, and E-SF increases when there are incomplete inputs (note that GE2E can only be applied when TI input is present).
Specifically, when TD signals are not available, \model\ outperforms GE2E, AF, SF, and E-SF by 47.7\%, 49.3\%, 48.3\%, and 47.0\% respectively, at a target FAR of 12.5\% (in Week~1).
This shows that our approach of inferring missing differential embeddings prior to system fusion is effective.

Fourth, we observe that E-SF fails to outperform SF consistently and they achieve similar FRR performances in most comparisons. This shows the disadvantage of score-level fusions, as all information has been already integrated into a single score and it becomes too hard to extract sufficient information from single scores to recover the missing scores from other systems.

We also separately confirmed the intuition that generating negative samples based on same-gender pairings improves model performance.
There is a relative 6\% FRR reduction when gender information is used for ruling out easy, different-gender samples.

\section{Conclusions}
\label{sec:conclusion}


We presented \model, a fusion architecture that combines predictions from both TD and TI speaker recognition systems at the embedding-level, by combining differential embeddings from both subsystems.
Special care is taken to allow the system to work when either TD or TI input is not available, as may happen due to the operational characteristics of voice assistants
(such as missing wakewords).
This is accomplished by estimating the missing differential embeddings from the embeddings that are available.
\model\ does not require fine tuning of the previously trained TI and TD models; both models can be directly used as is.
Experiments on Alexa voice assistant traffic data demonstrate that \model\ is substantially more effective than four baselines, i.e., a single TI GE2E model, a score-averaging fusion method, and two score-based neural network fusion methods.

\clearpage
\newpage
{
\bibliographystyle{IEEEtran}
\bibliography{refs}
}
\end{document}